\documentclass[10pt,twocolumn,letterpaper]{article}

\usepackage{cvpr}
\usepackage{times}
\usepackage{epsfig}
\usepackage{graphicx}
\usepackage{amsmath}
\usepackage{amssymb}

\usepackage{epsfig}
\usepackage{amsmath}
\usepackage{amssymb}
\usepackage{color}
\usepackage{comment}
\usepackage{algorithm}
\usepackage{algorithmic}
\usepackage{multirow}
\usepackage[pagebackref=true,breaklinks=true,letterpaper=true,colorlinks,bookmarks=false]{hyperref}

\cvprfinalcopy % *** Uncomment this line for the final submission

\def\cvprPaperID{2632} % *** Enter the CVPR Paper ID here

\ifcvprfinal\fi
\begin{document}

%%%%%%%%% TITLE
\title{Explaining Neural Networks Semantically and Quantitatively}

\author{Runjin Chen$^{a}$, Hao Chen$^{b}$, Ge Huang$^{a}$, Jie Ren$^{a}$, and Quanshi Zhang$^{a,*}$\\
$^{a}$John Hopcroft Center, Shanghai Jiao Tong University\\
$^{b}$Huazhong University of Science and Technology\\
}

\maketitle
%\thispagestyle{empty}

%%%%%%%%% ABSTRACT
\begin{abstract}
This paper\footnote[1]{Quanshi Zhang is the corresponding author. Runjin Chen and Hao Chen contribute equally.} presents a method to explain the knowledge encoded in a convolutional neural network (CNN) quantitatively and semantically. The analysis of the specific rationale of each prediction made by the CNN presents a key issue of understanding neural networks, but it is also of significant practical values in certain applications. In this study, we propose to distill knowledge from the CNN into an explainable additive model, so that we can use the explainable model to provide a quantitative explanation for the CNN prediction. We analyze the typical bias-interpreting problem of the explainable model and develop prior losses to guide the learning of the explainable additive model. Experimental results have demonstrated the effectiveness of our method.
\end{abstract}

%%%%%%%%% BODY TEXT
\section{Introduction}

Convolutional neural networks (CNNs)~\cite{CNN,CNNImageNet,ResNet} have achieved superior performance in various tasks. Besides the discrimination power of neural networks, the interpretability of networks has received an increasing attention in recent years.

\textbf{Motivation, trustiness of CNNs:} The network interpretability is directly related to the trustiness of a CNN, which is crucial in critical applications. As mentioned in \cite{CNNBias}, a high testing accuracy cannot fully ensure correct logics in neural networks, owing to the potential bias in datasets and feature representations. Instead, it is common for a CNN to use unreliable reasons for prediction.

Traditional studies usually interpreted neural networks at the pixel level, such as the visualization of network features~\cite{CNNVisualization_1,CNNVisualization_2,CNNVisualization_3,FeaVisual,visualCNN_grad,visualCNN_grad_2}, the extraction of pixel-level correlations between network inputs and outputs~\cite{CNNInfluence,trust,shap}.

\begin{figure}
\centering
\includegraphics[width=0.99\linewidth]{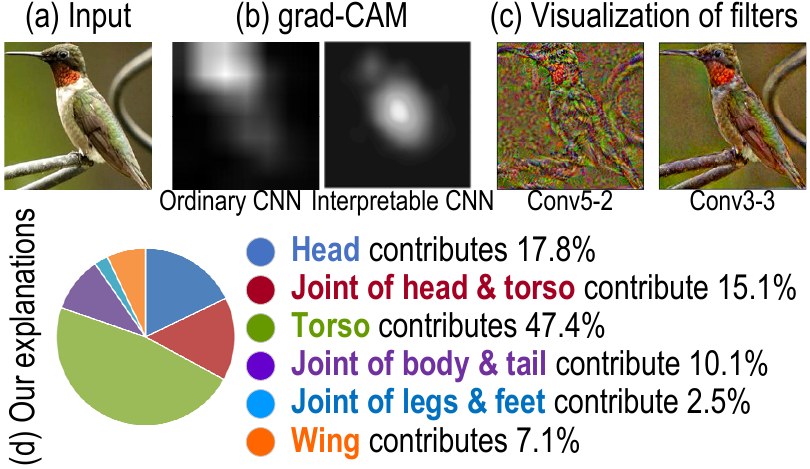}
\vspace{2pt}
\caption{Different types of explanations for CNNs. We compare (d) our task of quantitatively and semantically explaining CNN predictions with previous studies of interpreting CNNs, such as (b) the grad-CAM~\cite{visualCNN_grad_2} and (c) CNN visualization~\cite{CNNVisualization_2}. Given an input image (a) Our method generates a report, which quantitatively explains which object parts activate the CNN and how much these parts contribute to the prediction.}
\label{fig:task}
\end{figure}

In contrast to above qualitative analysis of CNNs, our semantically and quantitatively clarifying the logic of each network prediction is a more trustworthy way to diagnose feature representations of neural networks. Fig.~\ref{fig:task} compares our explanation with previous studies.

\begin{figure*}
\centering
\includegraphics[width=0.9\linewidth]{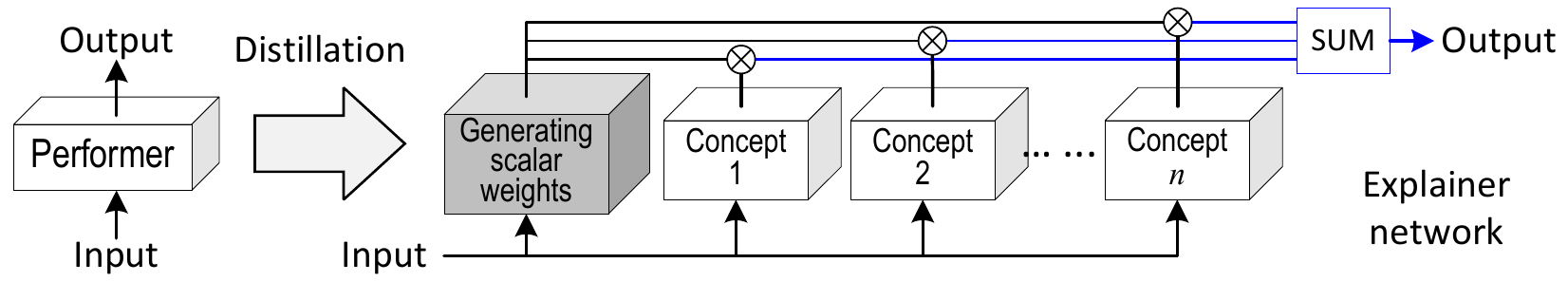}
\vspace{2pt}
\caption{Task. We distill knowledge of a performer into an explainer as a paraphrase of the performer's representations. The explainer decomposes the output score into value components of semantic concepts, thereby obtaining semantic explanations for the performer.}
\label{fig:model}
\end{figure*}

\noindent
$\bullet\;$\textit{Semantic explanations:} We hope to explain the logic of each network prediction using clear visual concepts, instead of using middle-layer features without clear meanings or simply extracting pixel-level correlations between network inputs and outputs. Semantic explanations satisfy specific demands in real applications.

\noindent
$\bullet\;$\textit{Quantitative explanations:} In contrast to traditional qualitative explanations for neural networks~\cite{CNNVisualization_1,CNNVisualization_2,CNNVisualization_3,FeaVisual,visualCNN_grad,visualCNN_grad_2,CNNInfluence,trust,shap}, quantitative explanations enable people to diagnose feature representations inside neural networks and help neural networks earn trust from people. We expect the neural network to provide the quantitative rationale of the prediction, \emph{i.e.} clarifying which visual concepts activate the neural network and how much they contribute to the prediction score. Figs.~\ref{fig:VOC_contri} and \ref{fig:CelebA_contri} show our explanations for CNN predictions. Predictions whose explanations conflict with people's common sense reflect problematic feature representations inside the CNN.

However, above two requirements of ``semantic explanations'' and ``quantitative explanations'' present core challenges of understanding neural networks. To the best of our knowledge, no previous studies simultaneously explained network predictions using clear visual concepts and quantitatively decomposed the prediction score into value components of these visual concepts.

\textbf{Task:} In order to explain the specific rationale of each network prediction semantically and quantitatively, in this study, we propose to learn another neural network, namely an \textit{explainer} network. Accordingly, the target CNN is termed a \textit{performer} network. Besides the performer, we also require a set of models that are pre-trained to detect different visual concepts. These visual concepts will be used to explain the logic of the performer's prediction. We are also given input images of the performer without any additional annotations on the images.

The explainer is learned to uses the pre-trained visual concepts to mimic the logic inside the performer, \emph{i.e.} the explainer uses features of the visual concepts to generate similar prediction scores.

As shown in Fig.~\ref{fig:model}, the explainer is designed as an additive model, which decomposes the prediction score into the sum of multiple value components. Each value component is computed based on a specific visual concept. In this way, we can roughly consider these value components as quantitative contributions of the visual concepts to the final prediction score.

More specifically, we learn the explainer via knowledge distillation. We do \textbf{not} use any ground-truth annotations on input images to supervise the explainer. It is because the task of the explainer is not to achieve a high prediction accuracy, but to mimic the performer's logic in prediction, even when the performer's prediction is incorrect.

Thus, the explainer can be regarded as a semantic paraphrase of feature representations inside the performer, and we can use the explainer to understand the logic of the performer's prediction. Theoretically, the explainer cannot precisely recover the exact prediction score of the performer, owing to the limit of the representation capacity of visual concepts. The difference of the prediction score between the performer and the explainer corresponds to the information that cannot be explained by the visual concepts.

\textbf{Explaining black-box networks or learning interpretable networks:} Some recent studies~\cite{interpretableCNN,capsule} learned neural networks with interpretable middle-layer features. Interpretable neural networks usually have specific requirements for structures~\cite{capsule} or losses~\cite{interpretableCNN}, which limit the model flexibility and applicability. Meanwhile, the interpretability of features is not equivalent to, and usually even conflicts with the discrimination power of features~~\cite{interpretableCNN,capsule}.

%In comparisons, the explainer summarizes the rationale of each prediction of the performer at the semantic level and analyzes quantitative contributions of different visual concepts.

In comparisons, the explainer explains the performer without affecting the original discrimination power of the performer. Compared to forcing the performer to learn interpretable features, our strategy of explaining the performer solves the dilemma between the interpretability and the discriminability.

\textbf{Core challenges:} Distilling knowledge from a pre-trained neural network into an additive model usually suffers from the problem of bias-interpreting. When we use a large number of visual concepts to explain the logic inside the performer, the explainer may biasedly select very few visual concepts, instead of all visual concepts, as the rationale of the prediction (see Fig.~\ref{fig:distri}). Just like the typical over-fitting problem, theoretically, the bias interpreting is an ill-defined problem. Therefore, we propose new losses for prior weights of visual concepts to overcome the bias-interpreting problem. The prior weights push the explainer to compute a similar Jacobian of the prediction score \emph{w.r.t.} visual concepts as the performer in early epochs, in order to avoid bias-interpreting.

\textbf{Contributions} of this study are summarized as follows. (i) In this study, we focus on a new explanation strategy, \emph{i.e.} semantically and quantitatively explaining CNN predictions. (ii) We propose a new method to explain neural networks, \emph{i.e.} distilling knowledge from a pre-trained performer into an interpretable additive explainer. Our strategy of using the explainer to explain the performer avoids hurting the discrimination power of the performer. (iii) We develop novel losses to overcome the typical bias-interpreting problem. Preliminary experimental results have demonstrated the effectiveness of the proposed method. (iv) Theoretically, the proposed method is a generic solution to the problem of interpreting neural networks. We have applied our method to different benchmark CNNs for different applications, which has proved the broad applicability of our method.

\section{Related work}

In this paper, we limit our discussion within the scope of understanding feature representations of neural networks.

\textbf{Network visualization:} The visualization of feature representations inside a neural network is the most direct way of opening the black-box of the neural network. Related techniques include gradient-based visualization~\cite{CNNVisualization_1,CNNVisualization_2,CNNVisualization_3,CNNVisualization_6} and up-convolutional nets~\cite{FeaVisual} to invert feature maps of conv-layers into images. However, recent visualization results with clear semantic meanings were usually generated with strict constraints. These constraints made visualization results biased towards people's preferences. Subjectively visualizing all information of a filter usually produced chaotic results. Thus, there is still a considerable gap between network visualization and semantic explanations for neural networks.

\textbf{Network diagnosis:} Some studies diagnose feature representations inside a neural network. \cite{CNNAnalysis_2} measured features transferability in intermediate layers of a neural network. \cite{CNNVisualization_5} visualized feature distributions of different categories in the feature space. \cite{trust,shap,patternNet,visualCNN_grad,visualCNN_grad_2} extracted rough pixel-level correlations between network inputs and outputs, \emph{i.e.} estimating image regions that directly contribute the network output. Network-attack methods~\cite{CNNInfluence,CNNAnalysis_1} computed adversarial samples to diagnose a CNN. \cite{banditUnknown} discovered knowledge blind spots of a CNN in a weakly-supervised manner. \cite{CNNBias} examined representations of conv-layers and automatically discover biased representations of a CNN due to the dataset bias. However, above methods usually analyzed a neural network at the pixel level and did not summarize the network knowledge into clear visual concepts.

%\cite{explanatoryGraph} summarized the six types of semantics into ``parts'' and ``textures.''

\cite{Interpretability} defined six types of semantics for CNN filters, \emph{i.e.} objects, parts, scenes, textures, materials, and colors. Then, \cite{CNNSemanticDeep} proposed a method to compute the image-resolution receptive field of neural activations in a feature map. Fong and Vedaldi~\cite{net2vector} analyzed how multiple filters jointly represented a certain semantic concept. Other studies retrieved middle-layer features from CNNs representing clear concepts. \cite{ObjectDiscoveryCNN_2} retrieved features to describe objects from feature maps, respectively. \cite{CNNSemanticDeep,CNNSemanticDeep2} selected neural units to describe scenes. Note that strictly speaking, each CNN filter usually represents a mixture of multiple semantic concepts. Unlike previous studies, we are more interested in analyzing the quantitative contribution of each semantic concept to each prediction, which was not discussed in previous studies.

\textbf{Learning interpretable representations:} A new trend in the scope of network interpretability is to learn interpretable feature representations in neural networks~\cite{LogicRuleNetwork,CNNCompositionality,Parsimonious} in an un-/weakly-supervised manner. Capsule nets~\cite{capsule} and interpretable RCNN~\cite{InterRCNN} learned interpretable features in intermediate layers. InfoGAN~\cite{infoGAN} and $\beta$-VAE~\cite{betaVAE} learned well-disentangled codes for generative networks. Interpretable CNNs~\cite{interpretableCNN} learned filters in intermediate layers to represent object parts without given part annotations. However, as mentioned in \cite{Interpretability}, interpretable features usually do not have a high discrimination power. Therefore, we use the explainer to interpret the pre-trained performer without hurting the discriminability of the performer.

\textbf{Explaining neural networks via knowledge distillation:} Distilling knowledge from a black-box model into an explainable model is an emerging direction in recent years. In contrast, we pursue the explicitly quantitative explanation for each CNN prediction. \cite{additiveExplainer2} learned an explainable additive model, and \cite{additiveExplainer} distilled knowledge of a network into an additive model. \cite{distillDecisionTree,TreeDistill,TreeDistill2,RNNTree} distilled representations of neural networks into tree structures. These methods did not explain the network knowledge using human-interpretable semantic concepts. More crucially, compared to previous additive models~\cite{additiveExplainer}, our research successfully overcomes the bias-interpreting problem, which is the core challenge when there are lots of visual concepts for explanation.

%\cite{explanatoryTree_arXiv} used a tree structure to summarize the inaccurate rationale of each CNN prediction into generic decision-making models for a number of samples.

\section{Algorithm}

In this section, we distill knowledge from a pre-trained performer $f$ to an explainable additive model. We are given a performer $f$ and $n$ neural networks $\{f_{i}|i=1,2,\ldots,n\}$ that are pre-trained to detect $n$ different visual concepts. We learn the $n$ neural networks along with the performer, and the $n$ neural networks are expected to share low-layer features with the performer. Our method also requires a set of training samples for the performer $f$. The goal of the explainer is to use inference values of the $n$ visual concepts to explain prediction scores of the performer. Note that we do not need any annotations on training samples \emph{w.r.t.} the task, because additional supervision will push the explainer towards a good performance of the task, instead of objectively reflecting the knowledge in the performer.

\begin{figure*}
\centering
\includegraphics[width=0.9\linewidth]{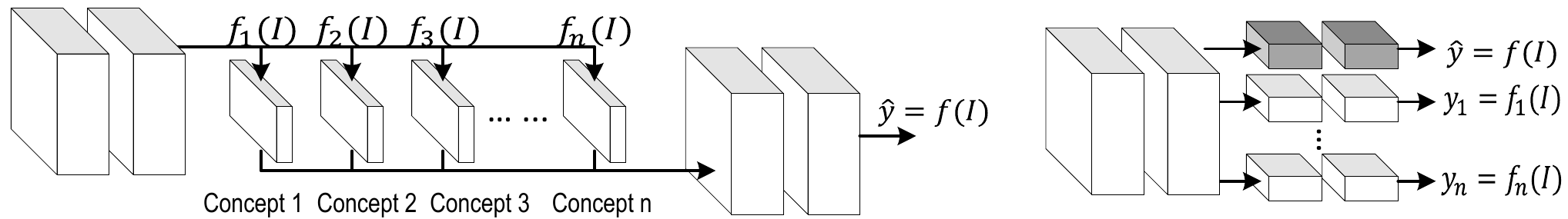}
\vspace{2pt}
\caption{Two typical types of neural networks. (left) A performer models interpretable visual concepts in its intermediate layers. For example, each filter in a certain conv-layer represents a specific visual concept. (right) The performer and visual concepts are jointly learned, and they share features in intermediate layers.}
\label{fig:case}
\end{figure*}

Given an input image $I$, let $\hat{y}=f(I)$ denote the output of the performer. Without loss of generality, we assume that $\hat{y}$ is a scalar. If the performer has multiple outputs (\emph{e.g.} a neural network for multi-category classification), we can learn an explainer to interpret each scalar output of the performer. In particular, when the performer takes a softmax layer as the last layer, we use the feature score before the softmax layer as $\hat{y}$, so that $\hat{y}$'s neighboring scores will not affect the value of $\hat{y}$.

We design the following additive explainer model, which uses a mixture of visual concepts to approximate the function of the performer. The explainer decomposes the prediction score $\hat{y}$ into value components of pre-defined visual concepts.
\begin{small}
\begin{equation}
\begin{split}
\hat{y}&\approx\!\!\!\!\underbrace{\alpha_{1}(I)\cdot y_{1}}_{\textrm{Quantitative contribution}\atop\textrm{from the first visual concept}}\!\!\!\!+\alpha_{2}(I)\cdot y_{2}+\ldots+\alpha_{n}(I)\cdot y_{n}+b,\\
y_{i}&=f_{i}(I),\quad i=1,2,\ldots,n\!\!
\end{split}
\end{equation}
\end{small}
where $y_{i}$ and $\alpha_{i}(I)$ denote the scalar value and the weight for the $i$-th visual concept, respectively. $b$ is a bias term. $y_{i}$ is given as the strength or confidence of the detection of the $i$-th visual concept. We can regard the value of $\alpha_{i}(I)\cdot y_{i}$ as the quantitative contribution of the $i$-th visual concept to the final prediction. In most cases, the explainer cannot recover all information of the performer. The prediction difference between the explainer and the performer reflects the limit of the representation capacity of visual concepts.

According to the above equation, the core task of the explainer is to estimate a set of weights ${\boldsymbol\alpha}=[\alpha_{1},\alpha_{2},\ldots,\alpha_{n}]$, which minimizes the difference of the prediction score between the performer and the explainer. Different input images may obtain different weights ${\boldsymbol\alpha}$, which correspond to different decision-making modes of the performer. For example, a performer may mainly use head patterns to classify a standing bird, while it may increase the weight for the wing concept to classify a flying bird. Therefore, we design another neural network $g$ with parameters ${\boldsymbol\theta}_{g}$ (\emph{i.e.} the explainer), which uses the input image $I$ to estimate the $n$ weights. We learn the explainer with the following knowledge-distillation loss.
\begin{small}
\begin{equation}
{\boldsymbol\alpha}=g(I),\qquad L=\Vert\hat{y}-\sum_{i=1}^{n}\alpha_{i}\cdot y_{i}-b\Vert^2
\end{equation}
\end{small}
However, without any prior knowledge about the distribution of the weight $\alpha_{i}$, the learning of $g$ usually suffers from the problem of bias-interpreting. The neural network $g$ may biasedly select very few visual concepts to approximate the performer as a shortcut solution, instead of sophisticatedly learning relationships between the performer output and all visual concepts.

Thus, to overcome the bias-interpreting problem, we use a loss $\mathcal{L}$ for priors of ${\boldsymbol\alpha}$ to guide the learning process in early epochs.
\begin{small}
\begin{equation}
\begin{split}
&\min_{{\boldsymbol\theta}_{g},b}Loss,\qquad Loss=L+\lambda(t)\cdot \mathcal{L}({\boldsymbol\alpha},{\bf w}),\\
&\textrm{s.t.} \lim_{t\rightarrow\infty}\lambda(t)=0
\end{split}
\end{equation}
\end{small}
where ${\bf w}$ denotes prior weights, which represent a rough relationship between the performer's prediction value and $n$ visual concepts. Just like ${\boldsymbol\alpha}$, different input images also have different prior weights ${\bf w}$. The loss $\mathcal{L}({\boldsymbol\alpha},{\bf w})$ penalizes the dissimilarity between ${\boldsymbol\alpha}$ and ${\bf w}$.

Note that the prior weights ${\bf w}$ are approximated with strong assumptions (we will introduce two different ways of computing ${\bf w}$ later). We use inaccurate ${\bf w}$ to avoid significant bias-interpreting, rather than pursue a high accuracy. Thus, we set a decreasing weight for $\mathcal{L}$, \emph{i.e.} $\lambda(t)=\frac{\beta}{\sqrt{t}}$, where $\beta$ is a scalar constant, and $t$ denotes the epoch number. In this way, we mainly apply the prior loss $\mathcal{L}$ in early epochs. Then, in late epochs, the influence of $\mathcal{L}$ gradually decreases, and our method gradually shifts its attention to the distillation loss for a high distillation accuracy.

We design two types of losses for prior weights, as follows.
\begin{small}
\begin{equation}
\!\mathcal{L}({\boldsymbol\alpha},{\bf w})\!=\!\left\{\!\begin{array}{ll}
crossEntropy(\frac{\boldsymbol\alpha}{\Vert{\boldsymbol\alpha}\Vert_1},\frac{\bf w}{\Vert{\bf w}\Vert_1}),&\!\!\!\forall i, \alpha_{i},w_{i}\geq 0\\
\Vert\frac{\boldsymbol\alpha}{\Vert{\boldsymbol\alpha}\Vert_2}-\frac{\bf w}{\Vert{\bf w}\Vert_2}\Vert_2^2, &\!\!\!\textrm{otherwise}
\end{array}
\right.\!\!\!
\label{eqn:prior}
\end{equation}
\end{small}
Some applications require a positive relationship between the prediction of the performer and each visual concept, \emph{i.e.} each weight $\alpha_{i}$ must be a positive scalar. In this case, we use the cross-entropy between ${\boldsymbol\alpha}$ and ${\bf w}$ as the prior loss. In other cases, the MSE loss between ${\boldsymbol\alpha}$ and ${\bf w}$ is used as the loss. $\Vert\cdot\Vert_1$ and $\Vert\cdot\Vert_2$ denote the L-1 norm and L-2 norm, respectively.

In particular, in order to ensure $\alpha_{i}\geq 0$ in certain applications, we add a non-linear activation layer as the last layer of $g$, \emph{i.e.} ${\boldsymbol\alpha}=\log[1+\exp(x)]$, where $x$ is the output of the last conv-layer.

\subsection{Computation of prior weights ${\bf w}$}
\label{sec:w}

In this subsection, we will introduce two techniques to efficiently compute rough prior weights ${\bf w}$, which are oriented to the following two cases in application.

\begin{figure*}
\centering
\includegraphics[width=0.95\linewidth]{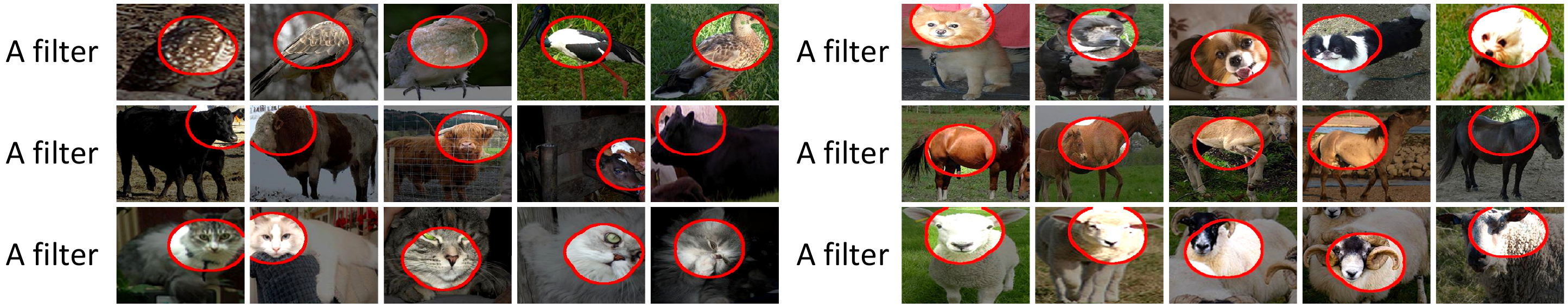}
\vspace{2pt}
\caption{We visualized interpretable filters in the top conv-layer of a CNN (Case 1), which were learned based on \cite{interpretableCNN}. We projected activation regions on the feature map of the filter onto the image plane for visualization. Each filter represented a specific object part through different images.}
\label{fig:part}
\end{figure*}

\textbf{Case 1, filters in intermediate conv-layers of the performer are interpretable:} As shown in Fig.~\ref{fig:case}(left), learning a neural network with interpretable filters is an emerging research direction in recent years. For example, \cite{interpretableCNN} proposed a method to learn CNNs for object classification, where each filter in a high conv-layer is exclusively triggered by the appearance of a specific object part. Fig.~\ref{fig:part} visualizes activation regions of these filters. Thus, we can interpret the classification score of an object as a linear combination of elementary scores for the detection of object parts. Because such interpretable filters are automatically learned without part annotations, the quantitative explanation for the CNN (\emph{i.e.} the performer) can be divided into the following two tasks: (i) annotating the name of the object part that is represented by each filter, and (ii) learning an explainer to disentangle the exact additive contribution of each filter (or each object part) to the performer output.

In this way, each $f_{i}$, $i=1,2,\ldots,n$, is given as an interpretable filter of the performer. According to \cite{CNNBias}, we can roughly represent the network prediction as
\begin{small}
\begin{equation}
\hat{y}\approx\sum_{i}w_{i}y_{i}+b,\quad\textrm{s.t.}\;\;\left\{\begin{array}{ll}
y_{i}&=\sum_{h,w}x_{hwi}\\
w_{i}&=\frac{1}{Z}\sum_{h,w}\frac{\partial\hat{y}}{\partial x_{hwi}}
\end{array}
\right.
\label{eqn:case1}
\end{equation}
\end{small}
where $x\in\mathbb{R}^{H\times W\times n}$ denotes a feature map of the interpretable conv-layer, and $x_{hwi}$ is referred to as the activation unit in the location $(h,w)$ of the $i$-th channel. $y_{i}$ measures the confidence of detecting the object part corresponding to the $i$-th filter. Here, we can roughly use the Jacobian of the network output \emph{w.r.t.} the filter to approximate the weight $w_{i}$ of the filter. $Z$ is for normalization. Considering that the normalization operation in Equation~(\ref{eqn:prior}) eliminates $Z$, we can directly use $\sum_{h,w}\frac{\partial \hat{y}}{\partial x_{hwi}}$ as prior weights ${\bf w}$ in Equation~(\ref{eqn:prior}) without a need to compute the exact value of $Z$.

\textbf{Case 2, neural networks for visual concepts share features in intermediate layers with the performer:} As shown in Fig.~\ref{fig:case}(right), given a neural network for the detection of multiple visual concepts, using certain visual concepts to explain a new visual concept is a generic way to interpret network predictions with broad applicability. Let us take the detection of a certain visual concept as the target $\hat{y}$ and use other visual concepts as $\{y_{i}\}$ to explain $\hat{y}$. All visual concepts share features in intermediate layers.

\begin{figure*}
\centering
\includegraphics[width=0.95\linewidth]{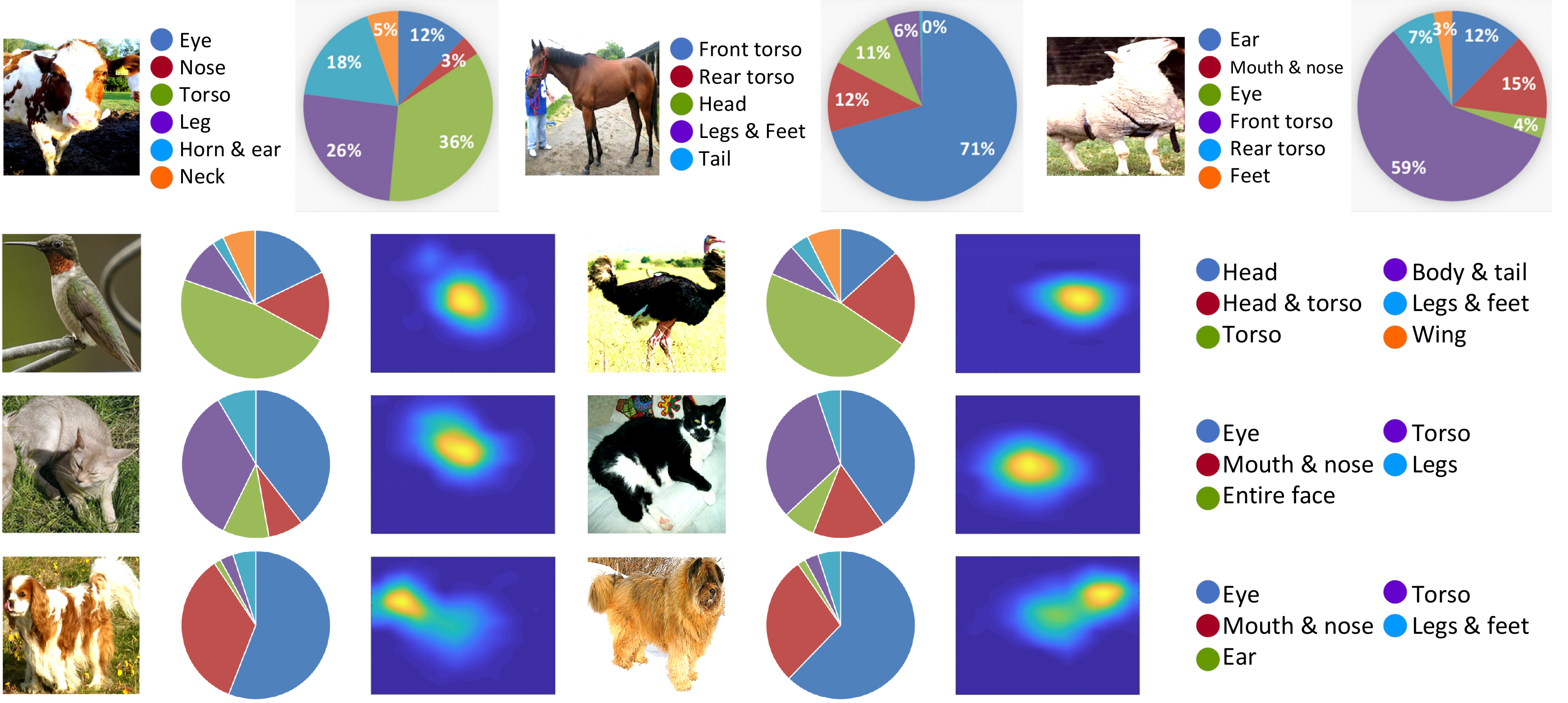}
\vspace{2pt}
\caption{Quantitative explanations for the object classification made by performers. We annotated the part that was represented by each interpretable filter in the performer, and we assigned contributions of filters $\alpha_{i}y_{i}$ to object parts. Thus, each pie chart illustrates contributions of different object parts for a specific input image. All object parts made positive contributions to the classification score. Heatmaps correspond to the grad-CAM visualization~\cite{visualCNN_grad_2} of feature maps of the performer to demonstrate the correctness of our explanations. Please see supplementary materials for more results.}
\label{fig:VOC_contri}
\end{figure*}

Then, we estimate a rough numerical relationship between $\hat{y}$ and the score of each visual concept $y_{i}$. Let $x$ be a middle-layer feature shared by both the target and the $i$-th visual concept. When we modify the feature $x$, we can represent the value change of $y_{i}$ using a Taylor series, $\Delta y_{i}=\frac{\partial y_{i}}{\partial x}\otimes\Delta x+O(\Delta^2 x)$, where $\otimes$ denotes the convolution operation. Thus, when we push the feature towards the direction of boosting $y_{i}$, \emph{i.e.} $\Delta x=\epsilon\frac{\partial y_{i}}{\partial x}$ ($\epsilon$ is a small constant), the change of the $i$-th visual concept can be approximated as $\Delta y_{i}=\epsilon\Vert\frac{\partial y_{i}}{\partial x}\Vert_{F}^2$, where $\Vert\cdot\Vert_{F}$ denotes the Frobenius norm. Meanwhile, $\Delta x$ also affects the target concept by $\Delta\hat{y}=\epsilon\frac{\partial\hat{y}}{\partial x}\otimes\frac{\partial y_{i}}{\partial x}$. Thus, we can roughly estimate the weight as $w_{i}=\frac{\Delta\hat{y}}{\Delta y_{i}}$.

\section{Experiments}

We designed two experiments to use our explainers to interpret different benchmark CNNs oriented to two different applications, in order to demonstrate the broad applicability of our method. In the first experiment, we used the detection of object parts to explain the detection of the entire object. In the second experiment, we used various face attributes to explain the prediction of another face attribute. We evaluated explanations obtained by our method qualitatively and quantitatively.

%We visualize the explanation for the performer's predictions as a qualitative evaluation of our method. We also quantitatively evaluated the limit of the representation capacity when we used clear visual concepts to approximate the performer.

\subsection{Experiment 1: using object parts to explain object classification}

In this experiment, we used the method proposed in \cite{interpretableCNN} to learn a CNN, where each filter in the top conv-layer represents a specific object part. We followed exact experimental settings in \cite{interpretableCNN}, which used the Pascal-Part dataset~\cite{SemanticPart} to learn six CNNs for the six animal\footnote[2]{Previous studies~\cite{SemanticPart,interpretableCNN} usually selected animal categories to test part localization, because animals usually contain non-rigid parts, which present significant challenges for part localization.} categories in the dataset. Each CNN was learned to classify the target animal from random images. We considered each CNN as a performer and regarded its interpretable filters in the top conv-layer as visual concepts to interpret the classification score.

In addition, when the CNN had been learned, we further annotated the object-part name corresponding to each filter based on visualization results (see Fig.~\ref{fig:part} for examples). We simply annotated each filter of the top conv-layer in a performer once, so the total annotation cost was $O(N)$, where $N$ is the filter number. Consequently, we assigned contributions of filters to its corresponding part, \emph{i.e.} {\small$Contri_{p}=\sum_{i\in\Omega_{p}}\alpha_{i}y_{i}$}, where $\Omega_{p}$ denotes the set of filter indexes that were assigned to the part $p$.

\textbf{Four types of CNNs as performers:} Following experimental settings in \cite{interpretableCNN}, we applied our method to four types of CNNs, including the AlexNet~\cite{CNNImageNet}, the VGG-M, VGG-S, and VGG-16 networks~\cite{VGG}, \emph{i.e.} we learned CNNs for six categories based on each network structure. Note that as discussed in \cite{interpretableCNN}, skip connections in residual networks~\cite{ResNet} increased the difficulty of learning part features, so they did not learn interpretable filters in residual networks.

\textbf{Learning the explainer:} The AlexNet/VGG-M/VGG-S/VGG-16 performer had 256/512/512/512 filters in its top conv-layer, so we set $n=256,512,512,512$ for these networks. We used the masked output of the top conv-layer as $x$ and plugged $x$ to Equation~(\ref{eqn:case1}) to compute $\{y_{i}\}$\footnote[3]{Please see supplementary materials for details}. We used the 152-layer ResNet~\cite{ResNet}\footnote[4]{Considering the small size of the input feature map, we removed the first max-pooling layer and the last average-pooling layer.} as $g$ to estimate weights of visual concepts\footnote[5]{Note that the input of the ResNet was the feature map of the top conv-layer, rather than the original image $I$ in experiments, so $g$ can be considered as a cascade of conv-layers in the AlexNet/VGGs and the ResNet.}. We set $\beta=10$ for the learning of all explainers. Note that all interpretable filters in the performer represented object parts of the target category on positive images, instead of describing random (negative) images. Intuitively, we needed to ensure a positive relationship between $\hat{y}$ and $y_{i}$. Thus, we filtered out negative prior weights $w_{i}\leftarrow\max\{w_{i},0\}$ and applied the cross-entropy loss in Equation~(\ref{eqn:prior}) to learn the explainer.

\begin{figure*}
\centering
\includegraphics[width=0.85\linewidth]{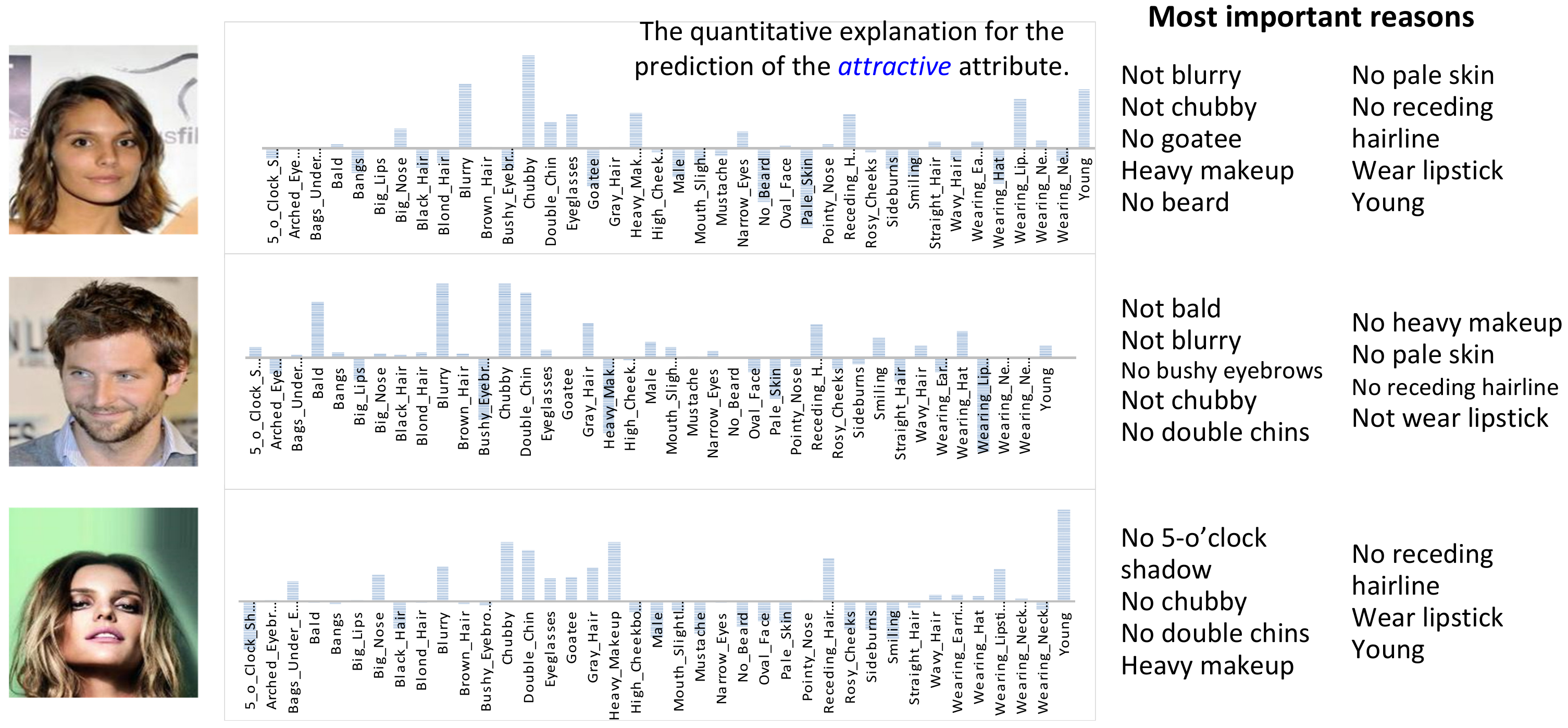}
\caption{Quantitative explanations for face-attribution predictions made by performers. Bars indicate elementary contributions $\alpha_{i}y_{i}$ from features of different face attributes, rather than prediction values $y_{i}$ of these attributes. For example, the network predicts a negative \textit{goatee} attribute $y_{\textrm{goatee}}<0$, and this information makes a positive contribution to the target \textit{attractive} attribute, $\alpha_{i}y_{i}>0$. Please see supplementary materials for more results.}
\label{fig:CelebA_contri}
\end{figure*}

\begin{figure*}
\centering
\includegraphics[width=0.99\linewidth]{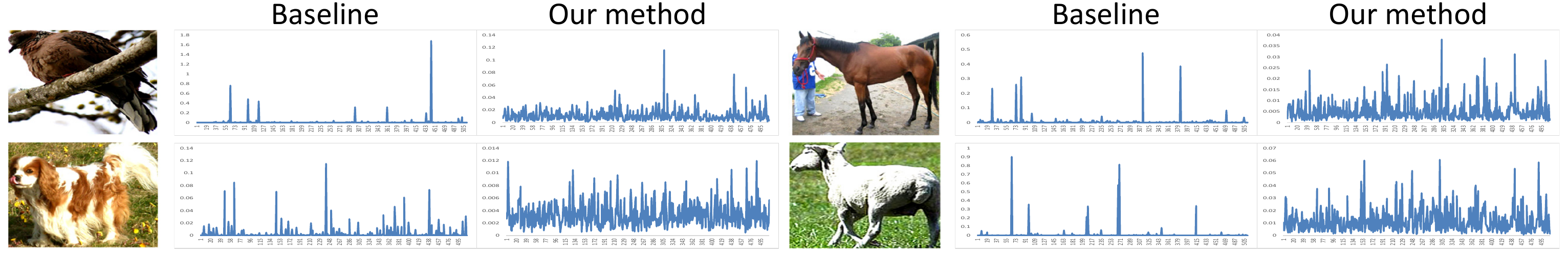}
\vspace{2pt}
\caption{We compared the contribution distribution of different visual concepts (filters) that was estimated by our method and the distribution that was estimated by the baseline. The horizontal axis and the vertical axis denote the filter index and the contribution value, respectively. The baseline usually used very few visual concepts to make predictions, which was a typical case of bias-interpreting.}
\label{fig:distri}
\end{figure*}
%In comparisons, our method provided a much more reasonable contribution distribution of visual concepts.

\textbf{Evaluation metric:} The evaluation has two aspects. Firstly, we evaluated the correctness of the estimated explanation for the performer prediction. The first metric is the \textit{error of the estimated contributions}. The explainer estimated numerical contributions of different visual concepts to the CNN prediction. For example, in Experiment 1, our method estimated the contribution of each annotated semantic part $p$, {\small$Contri_{p}$}. The error of the estimated contribution is computed as {\small$\mathbb{E}_{I\in{\bf I}}[\vert Contri_{p}-y_{p}^{*}\vert]/\mathbb{E}_{I\in{\bf I}}[y]$}, where $y$ denotes the CNN output \emph{w.r.t.} the image $I$; {\small$y_{p}^{*}$} is given as the ground-truth contribution of the part $p$. Let {\small$\Delta y_{p}$} denote the score change of $y$, when we removed all neural activations of filters corresponding the part $p$. Then, we computed {\small$y_{p}^{*}=y\frac{\Delta y_{p}}{\sum_{p'}\Delta y_{p'}}$}. In our experiments, we used semantic part annotations of the dog category to compute errors of the estimated contributions. In addition, we normalized the absolute contribution from each visual concept as a distribution of contributions $c_{i}=\vert\alpha_{i}y_{i}\vert/\sum_{j}\vert\alpha_{j}y_{j}\vert$. The entropy of contribution distribution $H({\bf c})$ can be considered as an indirect metric for bias-interpreting, although it is not directly related to the ground-truth of explanations. A biased explainer usually used very few visual concepts, instead of using most visual concepts, to approximate the performer, which led to a low entropy $H({\bf c})$.

Besides the quantitative evaluation, we also showed example explanations of for a qualitative evaluation of explanations. As shown in Fig.~\ref{fig:VOC_contri}, we used grad-CAM visualization~\cite{visualCNN_grad_2} of feature maps to prove the correctness of our explanations.

Secondly, we also measured the performer information that could not be represented by the visual concepts, which was unavoidable. We proposed two metrics for evaluation. The first metric is the \textit{prediction accuracy}. We compared the prediction accuracy of the performer with the prediction accuracy of using the explainer's output $\sum_{i}\alpha_{i}y_{i}+b$. Another metric is the \textit{relative deviation}, which measures a normalized output difference between the performer and the explainer. The relative deviation of the image $I$ is normalized as $\vert\hat{y}_{I}-\sum_{i}\alpha_{I,i}y_{I,i}-b\vert/(\max_{I'\in{\bf I}}\hat{y}_{I'}-\min_{I'\in{\bf I}}\hat{y}_{I'})$, where $\hat{y}_{I'}$ denotes the performer's output for the image $I'$.

Considering the limited representation power of visual concepts, the relative deviation on an image reflected inference patterns, which were not modeled by the explainer. The average relative deviation over all images was reported to evaluate the overall representation power of visual concepts. Note that our objective was \textbf{not} to pursue an extremely low relative deviation, because the limit of the representation power is an objective existence.

\begin{table*}[t]
\centering
\resizebox{0.75\linewidth}{!}{
\begin{tabular}{c|cccc||ccccc}
\hline
& \multicolumn{4}{|c||}{Experiment 1} & \multicolumn{5}{|c}{Experiment 2}\\
& AlexNet & VGG-M & VGG-S & VGG-16 & attractive & makeup & male & young & Avg.\\
\hline
Baseline
& 3.840
& 5.018
& 4.078
& 4.392
& 3.158
& 3.098
& 3.177
& 3.136
& 3.142
\\
Ours
& {\bf5.136}
& {\bf5.810}
& {\bf5.831}
& {\bf5.932}
& {\bf3.225}
& {\bf3.205}
& {\bf3.280}
& {\bf3.192}
& {\bf3.201}
\\
\hline
\end{tabular}}
\vspace{2pt}
\caption{Entropy of contribution distributions estimated by the explainer. A lower entropy of contribution distributions reflects more significant bias-interpreting. Our method suffered much less from the bias-interpreting problem than the baseline. Please see supplementary materials for more results.}
\label{tab:entropy}
\end{table*}

\begin{table*}
\centering
\resizebox{0.9\linewidth}{!}{
\begin{tabular}{c|c|cccc||ccccc}
\hline
\multicolumn{2}{c|}{} & \multicolumn{4}{|c||}{Experiment 1} & \multicolumn{5}{|c}{Experiment 2}\\
\multicolumn{2}{c|}{} & AlexNet & VGG-M & VGG-S & VGG-16 & attractive & makeup & male & young & Avg.\\
\hline
\multirow{2}{*}{Classification accuracy}& Performer
& 93.9
& 94.2
& 95.5
& 95.4
& 81.5
& 92.3
& 98.7
& 88.3
& 90.2
\\
& Explainer
& 92.6
& 93.6
& 95.4
& 95.6
& 74.9
& 88.9
& 97.7
& 82.0
& 85.9
\\
\hline
\multirow{2}{*}{Relative deviation} & Performer
& 0
& 0
& 0
& 0
& 0
& 0
& 0
& 0
& 0
\\
& Explainer
& 0.046
& 0.039
& 0.039
& 0.041
& 0.096
& 0.064
& 0.052
& 0.088
& 0.075
\\
\hline
\end{tabular}}
\vspace{2pt}
\caption{Classification accuracy and relative deviations of the explainer and the performer. We used relative deviations and the decrease of the classification accuracy to measure the information that could not be explained by pre-defined visual concepts. Please see supplementary materials for more results.}
\label{tab:acc}
\end{table*}

\begin{table}[t]
\centering
\resizebox{0.99\linewidth}{!}{
\begin{tabular}{c|ccccc|c}
& eye & mouth \& nose & ear & torso & leg & Avg.\\
\hline
Baseline
&0.399
&0.267
&0.045
&0.027
&0.084
&0.164
\\
Ours
&{\bf0.181}
&{\bf0.153}
&{\bf0.037}
&{\bf0.019}
&{\bf0.053}
&{\bf0.089}
\\
\hline
\end{tabular}}
\vspace{2pt}
\caption{Errors of the estimated object-part contributions. A lower error of our method indicates that the explanation yielded by our approach better fit the ground-truth rationale of a CNN prediction than the baseline.}
\label{tab:contri}
\end{table}

\subsection{Experiment 2: explaining face attributes based on face attributes}

In this experiment, we learned a CNN based on the VGG-16 structure to estimate face attributes. We used the Large-scale CelebFaces Attributes (CelebA) dataset~\cite{CelebA} to train a CNN to estimate 40 face attributes. We selected a certain attribute as the target and used its prediction score as $\hat{y}$. Other 39 attributes were taken as visual concepts to explain the score of $\hat{y}$ ($n=39$). The target attribute was selected from those representing global features of the face, \emph{i.e.} \textit{attractive}, \textit{heavy makeup}, \textit{male}, and \textit{young}. It is because global features can usually be described by local visual concepts, but the inverse is not. We learned an explainer for each target attribute. We used the same 152-layer ResNet structure as in Experiment 1 (expect for $n=39$) as $g$ to estimate weights. We followed the Case-2 implementation in Section~\ref{sec:w} to compute prior weights ${\bf w}$, in which we used the 4096-dimensional output of the first fully-connected layer as the shared feature $x$. We set $\beta=0.2$ and used the L-2 norm loss in Equation~(\ref{eqn:prior}) to learn all explainers. We used the same evaluation metric as in Experiment 1.

\subsection{Experimental results and analysis}

We compared our method with the traditional baseline of only using the distillation loss to learn the explainer. Tables~\ref{tab:contri} and \ref{tab:entropy} evaluate bias-interpreting of explainers that were learned using our method and the baseline. In particular, Fig.~\ref{fig:distri} illustrates the distribution of contributions of visual concepts $\{c_{i}\}$ when we learned the explainer using different methods. Our method suffered much less from the bias-interpreting problem than the baseline. According to Tables~\ref{tab:contri} and \ref{tab:entropy}, our method generated more informative explanations than the baseline. More crucially, part contributions estimated by our method better fit the ground truth than those estimated by the baseline. In contrast, the distillation baseline usually used very few visual concepts for explanation and ignored most strongly activated interpretable filters, which could be considered as bias-interpreting.

\begin{figure}
\centering
\includegraphics[width=0.99\linewidth]{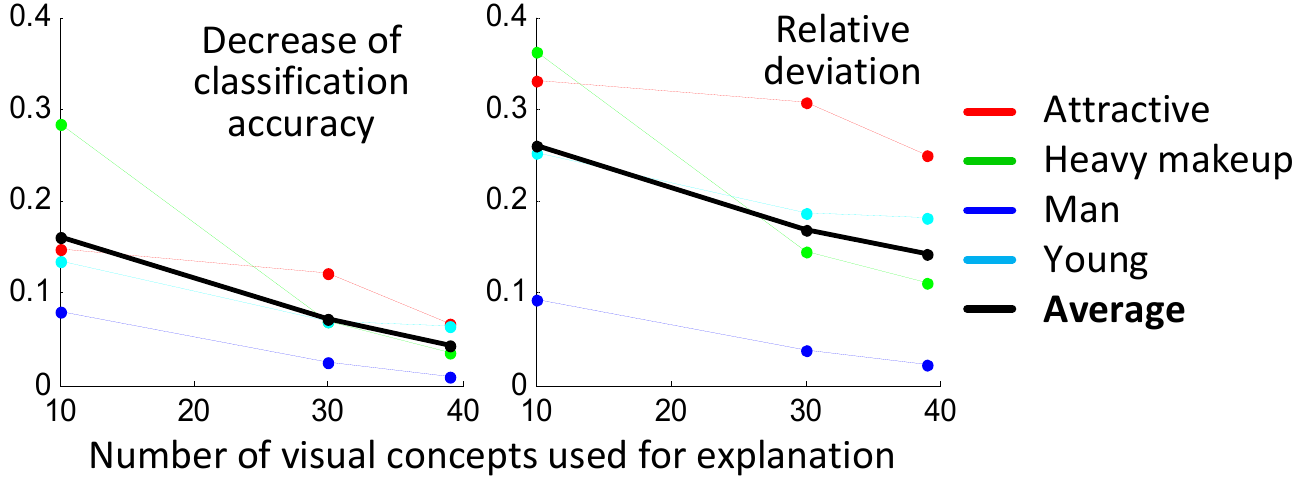}
\vspace{2pt}
\caption{Explanation capacity of using different numbers of visual concepts for explanation. We used the relative deviation and the decrease of the classification error (\emph{i.e.} the performer's accuracy minus the explainer's accuracy) using the explainer to roughly measure the limit of the explanation capacity. Using more visual concepts will increase the explanation capacity.}
\label{fig:curve}
\end{figure}

Figs.~\ref{fig:VOC_contri} and \ref{fig:CelebA_contri} show examples of quantitative explanations for the prediction made by the performer. In particular, we also used the grad-CAM visualization~\cite{visualCNN_grad_2} of feature maps of the performer to demonstrate the correctness of our explanations in Fig.~\ref{fig:VOC_contri}. In addition, Table~\ref{tab:acc} uses the classification accuracy and relative deviations of the explainer to measure the representation capacity of visual concepts.

\textbf{Selection of visual concepts:} How to select visual concepts for explanation is an important issue. The explanation capacity will decrease ff related concepts are not selected for explanation. Fig.~\ref{fig:curve} evaluates the change of the explanation capacity, when we randomly selected different numbers of visual concepts to learn explainers for CNN estimations of face attributes.

\section{Conclusion and discussions}

In this paper, we focus on a new explanation strategy, \emph{i.e.} explaining the logic of each CNN prediction semantically and quantitatively, which presents considerable challenges in the scope of understanding neural networks. We propose to distill knowledge from a pre-trained performer into an interpretable additive explainer. We can consider that the performer and the explainer encode similar knowledge. The additive explainer decomposes the prediction score of the performer into value components from semantic visual concepts, in order to compute quantitative contributions of different concepts. The strategy of using an explainer for explanation avoids decreasing the discrimination power of the performer. In preliminary experiments, we have applied our method to different benchmark CNN performers to prove the broad applicability.

Note that our objective is not to use pre-trained visual concepts to achieve super accuracy in classification/prediction. Instead, the explainer uses these visual concepts to mimic the logic of the performer as the performer. We evaluated the effects of selecting different numbers of visual concepts for explanation.

In particular, bias interpreting is the biggest challenge of using an additive explainer to interpret another neural network. In this study, we design two losses to overcome the bias-interpreting problems. Besides, in experiments, we measured the amount of the performer knowledge that could not be represented by the visual concepts in the explainer and used two metrics to evaluate the significance of bias interpreting.

{\small
\bibliographystyle{ieee}
\bibliography{TheBib}
}

\newpage
\onecolumn
\appendix
\section*{Detailed results}

\resizebox{0.6\linewidth}{!}{
\begin{tabular}{c|cc|cc|cc|cc}
\multicolumn{9}{}{Experiment 1}\\
\hline
\!\!\!& \multicolumn{2}{|c|}{\small AlexNet} & \multicolumn{2}{|c|}{\small VGG-M} & \multicolumn{2}{|c|}{\small VGG-S} & \multicolumn{2}{|c}{\small VGG-16}\\
\!\!\!&\!\!\! {\footnotesize Baseline} \!\!\!&\!\!\! {\small Ours} \!\!\!&\!\!\! {\footnotesize Baseline} \!\!\!&\!\!\! {\small Ours} \!\!\!&\!\!\! {\footnotesize Baseline} \!\!\!&\!\!\! {\small Ours} \!\!\!&\!\!\! {\footnotesize Baseline} \!\!\!&\!\!\! {\small Ours}\\
\hline
bird
& 3.851
& {\bf 5.129}
& 4.686
& {\bf 5.783}
& 4.177
& {\bf 5.799}
& 4.249
& {\bf 5.909}
\\
cat
& 3.458
& {\bf 5.056}
& 5.627
& {\bf 5.769}
& 4.099
& {\bf 5.824}
& 5.175
& {\bf 5.944}
\\
cow
& 3.895
& {\bf 5.178}
& 4.679
& {\bf 5.882}
& 4.027
& {\bf 5.832}
& 3.984
& {\bf 5.975}
\\
dog
& 3.934
& {\bf 5.163}
& 5.605
& {\bf 5.906}
& 4.150
& {\bf 5.888}
& 4.844
& {\bf 5.972}
\\
horse
& 4.032
& {\bf 5.194}
& 4.768
& {\bf 5.833}
& 3.860
& {\bf 5.881}
& 4.166
& {\bf 5.903}
\\
sheep
& 3.869
& {\bf 5.096}
& 4.745
& {\bf 5.689}
& 4.156
& {\bf 5.763}
& 3.936
& {\bf 5.889}
\\
\hline
Avg.
& 3.840
& {\bf 5.136}
& 5.018
& {\bf 5.810}
& 4.078
& {\bf 5.831}
& 4.392
& {\bf 5.932}
\\
\hline
\end{tabular}}

\vspace{10pt}
Entropy of contribution distributions. The entropy of contribution distributions reflects the level of bias-interpreting. The lower entropy indicates a larger bias. Our method suffered much less from the bias-interpreting problem than the baseline.

\resizebox{1.0\linewidth}{!}{
\begin{tabular}{c|c|cc||c|cc||c|cc||ccc}
\multicolumn{13}{}{Experiment 1}\\
\hline
\!\!\!& \multicolumn{3}{|c||}{\small AlexNet} & \multicolumn{3}{|c||}{\small VGG-M} & \multicolumn{3}{|c||}{\small VGG-S} & \multicolumn{3}{|c}{\small VGG-16}\\
\!\!\!&\!\!\! {\footnotesize Performer} \!\!\!&\!\!\! {\footnotesize Baseline} \!\!\!&\!\!\! {\small Ours} \!\!\!&\!\!\! {\footnotesize Performer} \!\!\!&\!\!\! {\footnotesize Baseline} \!\!\!&\!\!\! {\small Ours} \!\!\!&\!\!\! {\footnotesize Performer} \!\!\!&\!\!\! {\footnotesize Baseline} \!\!\!&\!\!\! {\small Ours} \!\!\!&\!\!\! {\footnotesize Performer} \!\!\!&\!\!\! {\footnotesize Baseline} \!\!\!&\!\!\! {\small Ours}\\
\hline
bird
& 92.8
& 90.5
& 90.8

& 96.8
& 97.3
& 98.0

& 96.5
& 97.0
& 96.8

& 97.3
& 98.3
& 98.0
\\
cat
& 96.3
& 94.8
& 95.0

& 94.3
& 95.8
& 95.8

& 95.3
& 96.3
& 95.5

& 94.3
& 97.0
& 95.3
\\
cow
& 93.4
& 87.3
& 89.6

& 95.2
& 92.9
& 92.9

& 94.4
& 94.4
& 94.4

& 91.1
& 97.2
& 93.4
\\
dog
& 92.5
& 92.0
& 92.0

& 93.8
& 94.0
& 94.5

& 95.3
& 93
& 94.8

& 94.5
& 96.8
& 97.0
\\
horse
& 91.4
& 89.1
& 89.6

& 92.9
& 88.1
& 87.6

& 92.9
& 92.7
& 92.7

& 99.2
& 95.2
& 95.2
\\
sheep
& 97.2
& 95.7
& 98.5

& 92.2
& 92.7
& 92.7

& 98.5
& 98.5
& 98.2

& 96.0
& 95.7
& 94.7
\\
\hline
Average
& 93.9
& 91.6
& {\bf92.6}

& 94.2
& 93.5
& {\bf93.6}

& 95.5
& 95.3
& {\bf95.4}

& 95.4
& {\bf96.7}
& 95.6
\\
\hline
\end{tabular}}
\resizebox{0.55\linewidth}{!}{
\begin{tabular}{c|cccc|c}
\multicolumn{6}{}{Experiment 2}\\
\hline
& Attractive & Makeup & Male & Young & Avg.\\
\hline
Performer
& 81.5
& 92.3
& 98.7
& 88.3
& 90.2
\\
\hline
Explainer, baseline
& 74.0
& 88.8
& 97.6
& 81.9
& 85.6
\\
Explainer, ours
& {\bf74.9}
& {\bf88.9}
& {\bf97.7}
& {\bf82.0}
& {\bf85.9}
\\
\hline
\end{tabular}}

\vspace{10pt}
Classification accuracy of the explainer and the performer. We use the the classification accuracy to measure the information loss when using an explainer to interpret the performer. Note that the additional loss for bias-interpreting successfully overcame the bias-interpreting problem, but did not decrease the classification accuracy of the explainer. Another interesting finding of this research is that sometimes, the explainer even outperformed the performer in classification. A similar phenomenon has been reported in \cite{BornAgain}. A possible explanation for this phenomenon is given as follows. When the student network in knowledge distillation had sufficient representation power, the student network might learn better representations than the teacher network, because the distillation process removed abnormal middle-layer features corresponding to irregular samples and maintained common features, so as to boost the robustness of the student network.

\resizebox{0.8\linewidth}{!}{
\begin{tabular}{c|cc|cc|cc|cc}
\multicolumn{9}{}{Experiment 1}\\
\hline
\!\!\!& \multicolumn{2}{|c|}{\small AlexNet} & \multicolumn{2}{|c|}{\small VGG-M} & \multicolumn{2}{|c|}{\small VGG-S} & \multicolumn{2}{|c}{\small VGG-16}\\
\!\!\!&\!\!\! {\footnotesize Baseline} \!\!\!&\!\!\! {\small Ours} \!\!\!&\!\!\! {\footnotesize Baseline} \!\!\!&\!\!\! {\small Ours} \!\!\!&\!\!\! {\footnotesize Baseline} \!\!\!&\!\!\! {\small Ours} \!\!\!&\!\!\! {\footnotesize Baseline} \!\!\!&\!\!\! {\small Ours}\\
\hline
bird
& 0.055
& 0.050
& 0.034
& 0.033
& 0.036
& 0.034
& 0.039
& 0.032
\\
cat
& 0.048
& 0.043
& 0.033
& 0.026
& 0.043
& 0.037
& 0.027
& 0.050
\\
cow
& 0.056
& 0.052
& 0.067
& 0.060
& 0.054
& 0.046
& 0.050
& 0.055
\\
dog
& 0.047
& 0.043
& 0.027
& 0.027
& 0.053
& 0.044
& 0.028
& 0.032
\\
horse
& 0.043
& 0.044

& 0.057
& 0.049

& 0.046
& 0.039

& 0.036
& 0.035
\\
sheep
& 0.052
& 0.046

& 0.048
& 0.038

& 0.040
& 0.038

& 0.038
& 0.042
\\
\hline
Average
& 0.050
& 0.046

& 0.044
& 0.039

& 0.045
& 0.039

& 0.036
& 0.041
\\
\hline
\end{tabular}}

\vspace{10pt}
Relative deviations of the explainer. The additional loss for bias-interpreting successfully overcame the bias-interpreting problem and just increased a bit (ignorable) relative deviation of the explainer.

\section*{Image-specific explanations v.s. generic explanations}

\cite{explanatoryTree_arXiv} used a tree structure to summarize the inaccurate rationale of each CNN prediction into generic decision-making models for a number of samples. This method assumed the significance of a feature to be proportional to the Jacobian \emph{w.r.t.} the feature, which is quite problematic. This assumption is acceptable for \cite{explanatoryTree_arXiv}, because the objective of \cite{explanatoryTree_arXiv} is to learn a generic explanation for a group of samples, and the inaccuracy in the explanation for each specific sample does not significantly affect the accuracy of the generic explanation. In comparisons, our method focuses on the quantitative explanation for each specific sample, so we design an additive model to obtain more convincing explanations.

\section*{Visualization of bias-interpreting}

\includegraphics[width=0.75\linewidth]{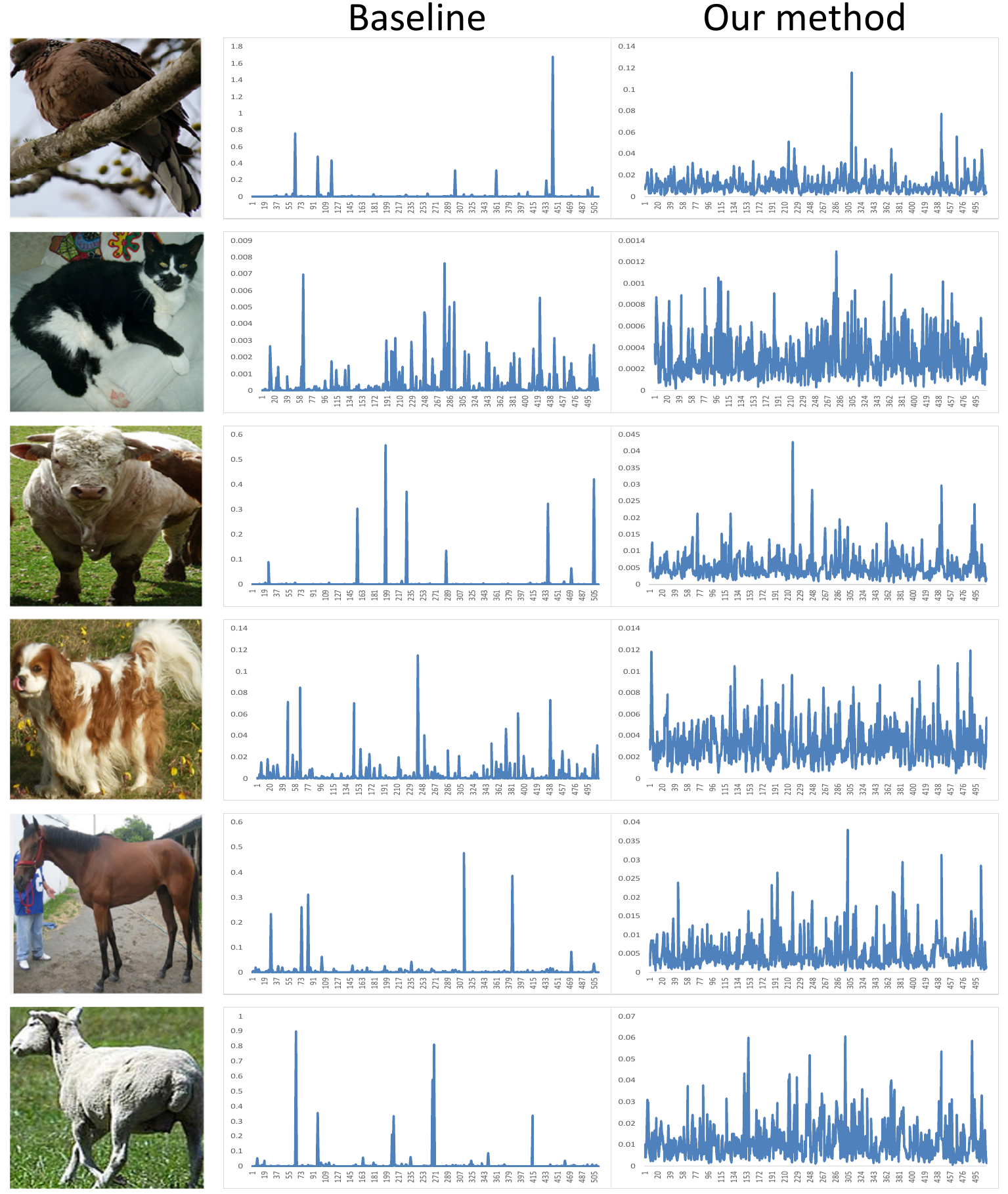}

We compared the contribution distribution of different visual concepts (filters) that was estimated by our method and the distribution that was estimated by the baseline. The baseline usually used very few visual concepts to make predictions, which was a typical case of bias-interpreting. In comparisons, our method provided a much more reasonable contribution distribution of visual concepts.

\newpage
\section*{Visualization of quantitative explanations}

\includegraphics[width=0.8\linewidth]{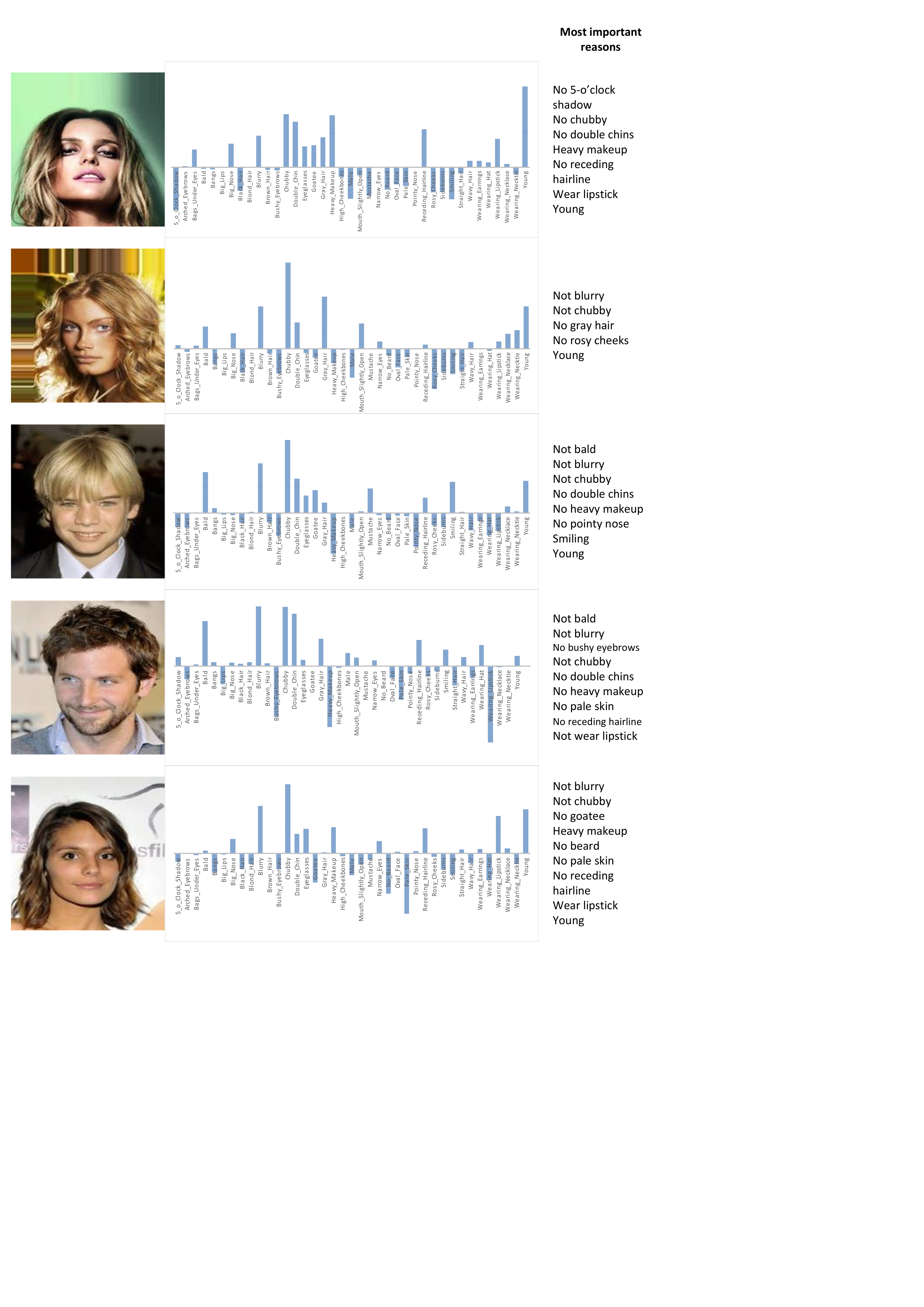}

Quantitative explanations for the \textit{attractive} attribute. Bars indicate elementary contributions $\alpha_{i}y_{i}$ from features of different face attributes, rather than the prediction of these attributes. For example, the network predicts a negative \textit{goatee} attribute $y_{\textrm{goatee}}<0$, and this information makes a positive contribution to the target \textit{attractive} attribute, $\alpha_{i}y_{i}>0$.

\newpage
\includegraphics[width=0.85\linewidth]{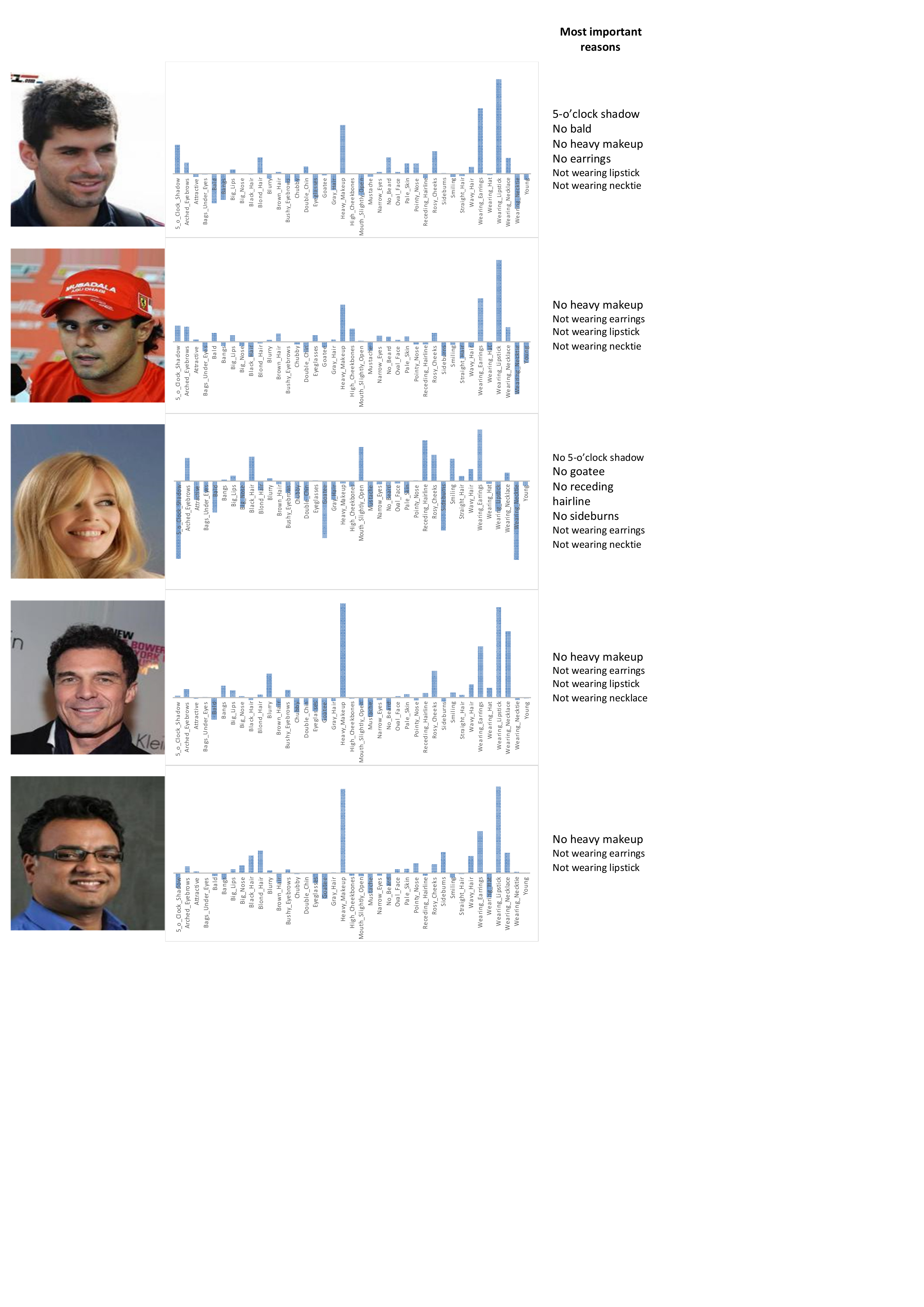}

Quantitative explanations $\alpha_{i}y_{i}$ for the \textit{male} attribute.

\newpage
\includegraphics[width=0.85\linewidth]{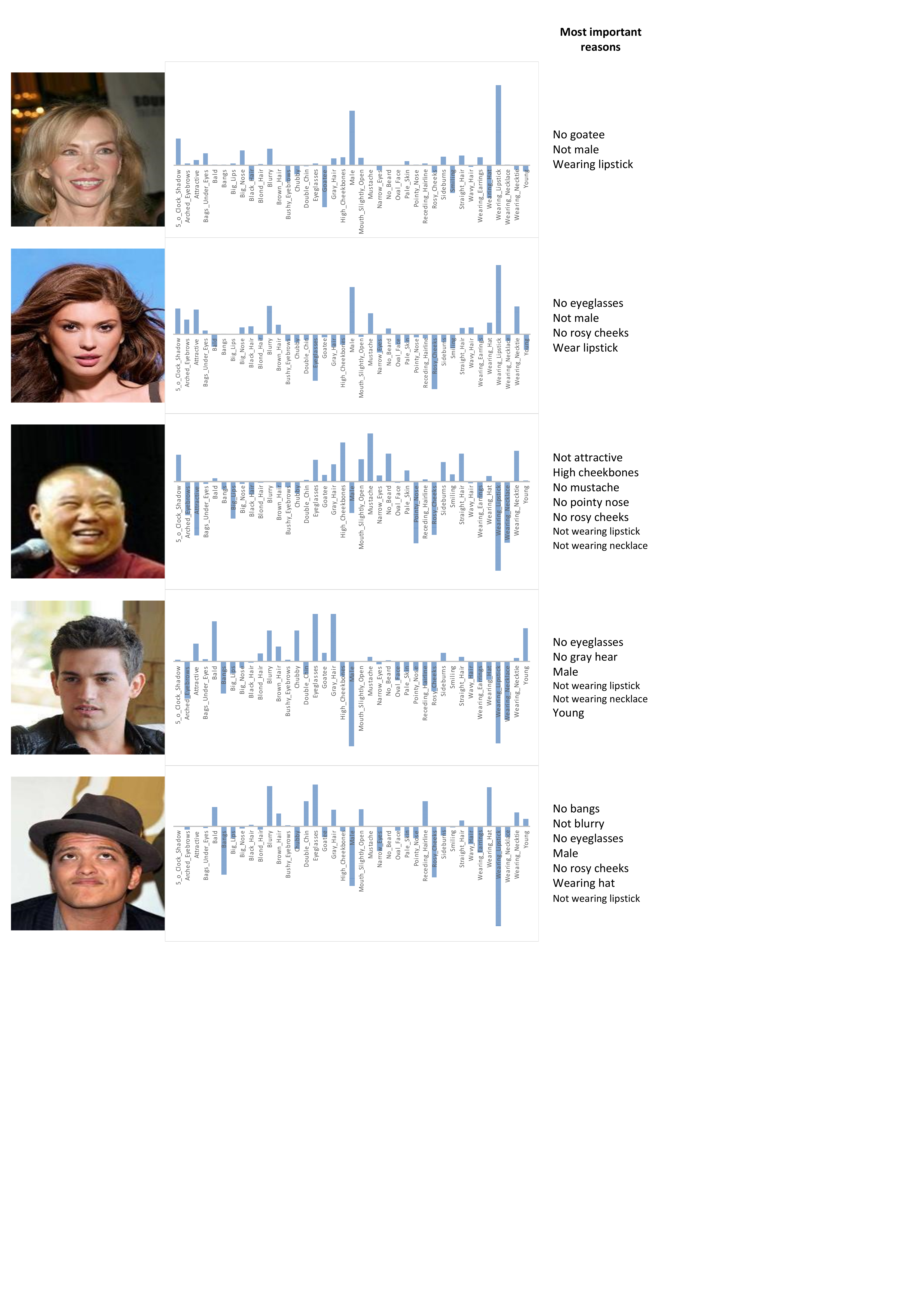}

Quantitative explanations $\alpha_{i}y_{i}$ for the \textit{heavy makeup} attribute.

\newpage
\includegraphics[width=0.85\linewidth]{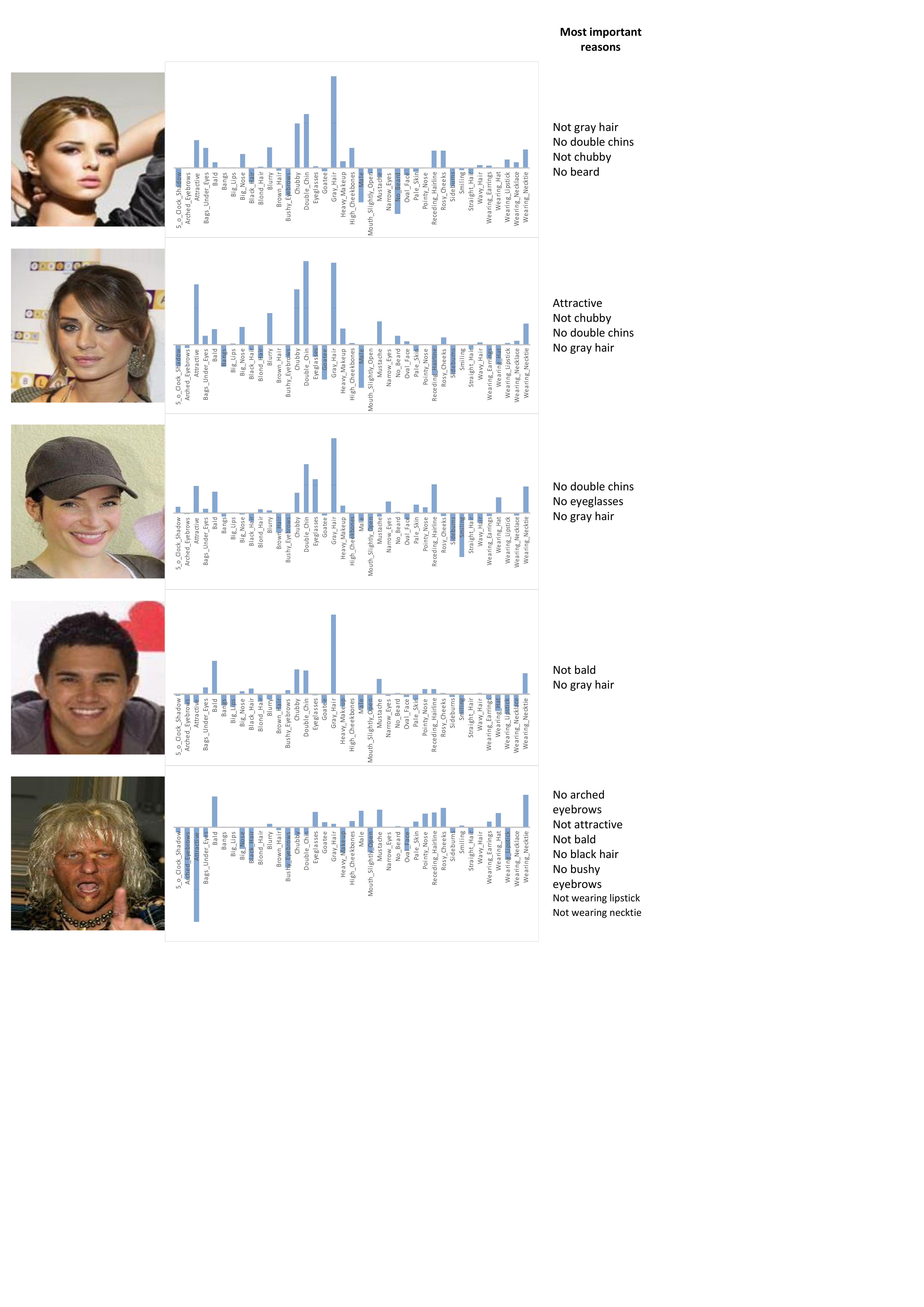}

Quantitative explanations $\alpha_{i}y_{i}$ for the \textit{young} attribute.

\newpage
\includegraphics[width=\linewidth]{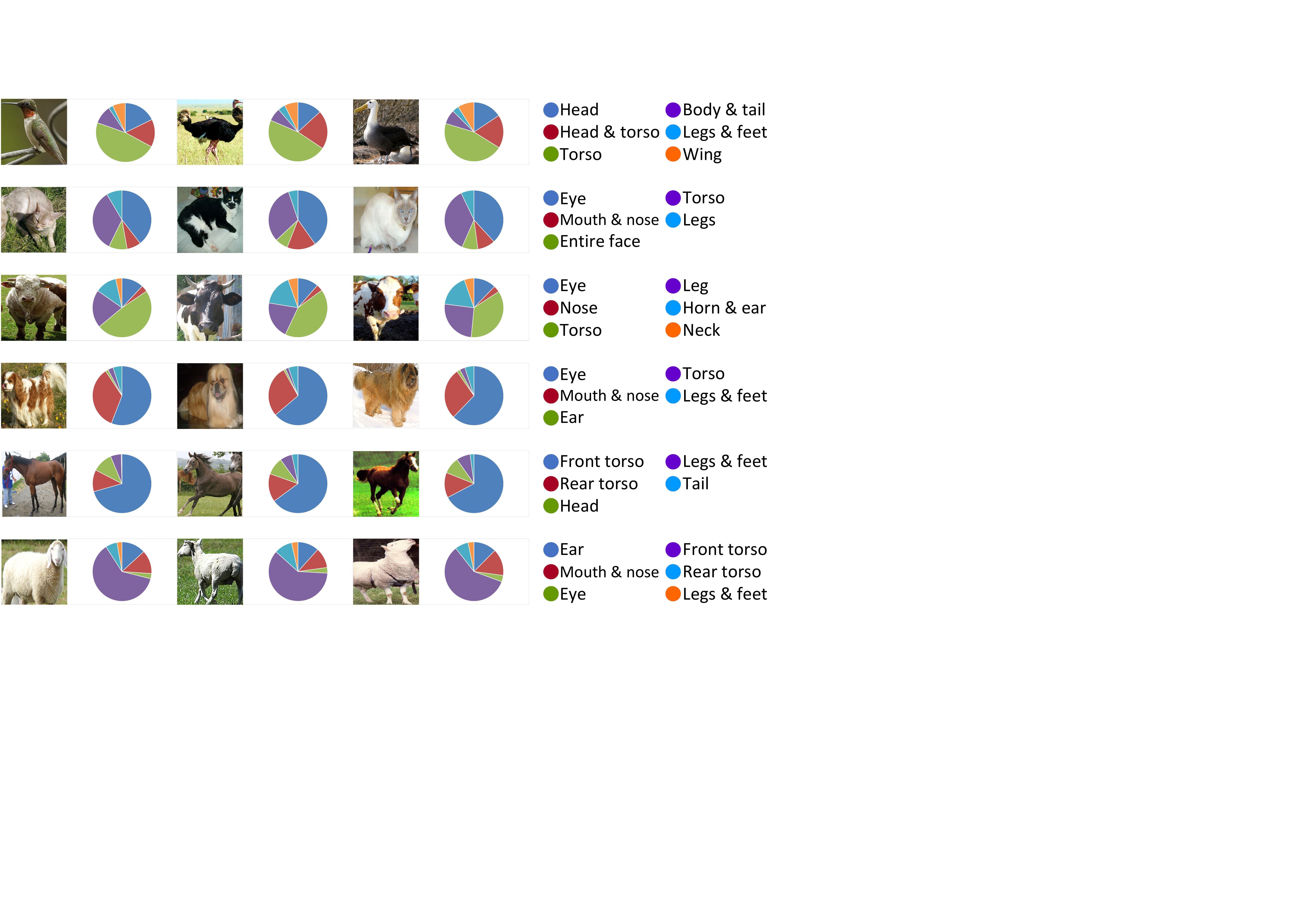}
\includegraphics[width=\linewidth]{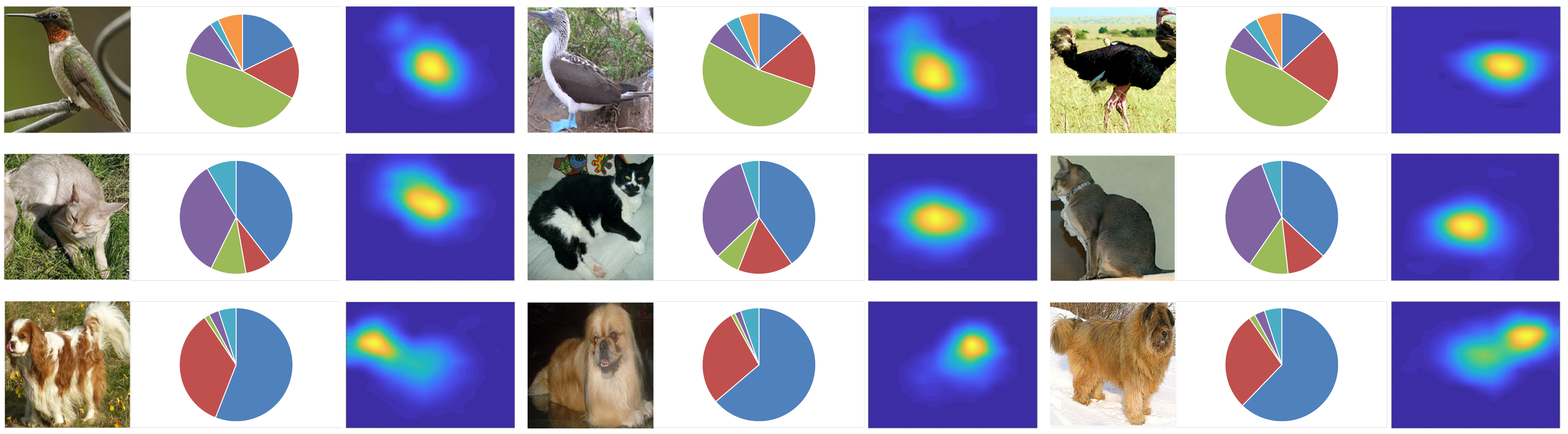}

Quantitative explanations for object classification. We assigned contributions of filters to their corresponding object parts, so that we obtained contributions of different object parts. According to top figures, we found that different images had similar explanations, \emph{i.e.} the CNN used similar object parts to classify objects. Therefore, we showed the grad-CAM visualization of feature maps~\cite{visualCNN_grad_2} on the bottom, which proved this finding.

\newpage
\section*{Visualization of interpretable filters}

\includegraphics[width=0.75\linewidth]{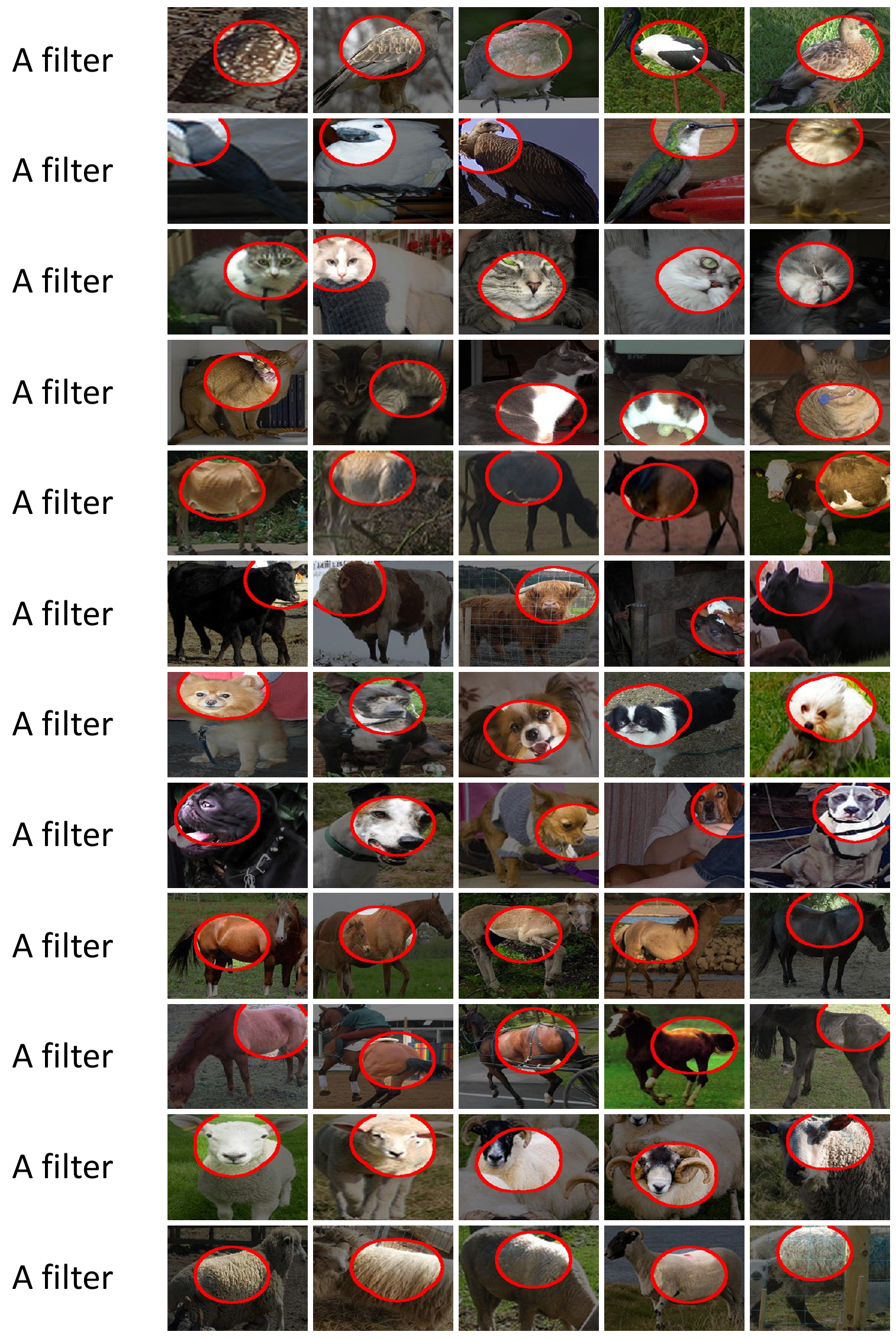}

We visualized interpretable filters in the top conv-layer of a CNN, which were learned based on \cite{interpretableCNN}. We projected activation regions on the feature map of the filter onto the image plane for visualization. Each filter represented a specific object part through different images.

\newpage
\section*{Details in Experiment 1}

We changed the order of the ReLU layer and the mask layer after the top conv-layer, \emph{i.e.} placing the mask layer between the ReLU layer and the top conv-layer. According to \cite{interpretableCNN}, this operation did not affect the performance of the pre-trained performer. We used the output of the mask layer as $x$ and plugged $x$ to the equation for Case 1 to compute $\{y_{i}\}$.

Because the distillation process did not use any ground-truth class labels, the explainer's output $\sum_{i}\alpha_{i}y_{i}+b$ was not sophisticatedly learned for classification. Thus, we used a threshold $\sum_{i}\alpha_{i}y_{i}+b>\tau$ ($\tau\approx0$), instead of 0, as the decision boundary for classification. $\tau$ was selected as the one that maximized the accuracy. Such experimental settings made a fairer comparison between the performer and the explainer.

\end{document}